\documentclass[11pt,letterpaper]{article}
\usepackage{naaclhlt2016}
\usepackage{times}
\usepackage{latexsym}
\usepackage{url}
\usepackage{graphicx}
\usepackage{booktabs,dcolumn}
\usepackage{multirow}
\usepackage{tabularx}
\usepackage{caption}
\usepackage{blindtext}

\usepackage{mylingmacros}
\usepackage{url}

\def\nl#1{\textit{#1}}

\naaclfinalcopy 

\newcommand{\wikibio}{\textsc{\scriptsize WikiBio}}
\newcommand{\wikibioastro}{\textsc{\scriptsize WikiBioAstro}}

\newcommand{\rnnlgHotel}{\textsc{\scriptsize RNNLG$_{Hotel}$}}
\newcommand{\rnnlgRestaurant}{\textsc{\scriptsize RNNLG$_{Restaurant}$}}
\newcommand{\rnnlgLaptop}{\textsc{\scriptsize RNNLG$_{Laptop}$}}
\newcommand{\rnnTV}{\textsc{\scriptsize RNNLG$_{TV}$}}
\newcommand{\rnn}{\textsc{\scriptsize RNNLG}}
\newcommand{\analogue}{\textsc{\scriptsize ImageDesc}}
\newcommand{\gmb}{\textsc{\scriptsize GMB}}

\newcommand{\MRInstances}{\scriptsize MR Inst.}
\newcommand{\OntComplexity}{\scriptsize Nb. of Attr.}
\newcommand{\size}{\scriptsize Size}

\newcolumntype{.}{D{.}{.}{-1}}

\title{Analysing Data-To-Text Generation Benchmarks}

\author{Laura Perez-Beltrachini \\ School of Informatics, University of Edinburgh \\  10 Crichton Street, Edinburgh EH8 9AB \\ Scotland\\
         \And 
         Claire Gardent \\ CNRS, LORIA, UMR 7503 \\ Vanoeuvre-l\`{e}s-Nancy, F-54506 \\ France}

\date{}

\begin{document}
\maketitle
\begin{abstract}

Recently, several data-sets associating data to text have been created
to train data-to-text surface realisers.  It is unclear however to
what extent the surface realisation task exercised by these data-sets
is linguistically challenging. Do these data-sets provide enough
variety to encourage the development of generic, high-quality
data-to-text surface realisers ?  In this paper, we argue that
these data-sets have important drawbacks. We back up our claim using
statistics, metrics and manual evaluation. We conclude by eliciting a
set of criteria for the creation of a data-to-text benchmark which
could help better support the development, evaluation and comparison
of linguistically sophisticated data-to-text surface
realisers.
\end{abstract}

\section{Introduction}

Recently, several data-sets associating data to text have been
constructed and used to train, mostly neural, data-to-text surface
realisers.  \cite{lebret-grangier-auli:2016:EMNLP2016} built a
biography data-set using Wikipedia articles and
infoboxes. \cite{wen2016multi} crowd sourced text for dialogue acts
bearing on multiple domains (restaurant, laptop, car and TV).
\cite{novikova-lemon-rieser:2016:INLG} present a data-set created by
crowd sourcing text and paraphrases from image illustrated frames.

In this paper, we examine the adequacy of these data-sets for
training high precision, wide coverage and linguistically
sophisticated, surface realisers. We focus on the following questions:
\\

\noindent
\textit{Lexical richness:} Is the data-set lexically varied ? Domain specific data may lead to highly repetitive text structures due to a small number of lexical items. \\
\textit{Syntactic variation:} is the data-set syntactically varied and in particular, does it include text of varied syntactic \textit{complexity}   ?\\
\textit{Semantic adequacy:} for each data-text pair contained in the data-set, does the text match the information contained in the data ?\\
\textit{Linguistic adequacy:} Language provides many ways of expressing the same content. How much do the available data-sets support the learning of paraphrases ?

Using existing tools and metrics, we measure and compare the syntactic
complexity and lexical richness of the three data-sets mentioned
above. We provide some comparative statistics for size, number of
distinct attributes and number of distinct data units. And we report
on a human evaluation of their semantic adequacy (data/text match
precision). Our analysis reveals different weak points for each of the
data-sets. We conclude by eliciting a number of important criteria that
should help promote the development of a higher quality data-to-text
benchmarks for the training of linguistically sophisticated surface
realisers.

\section{Datasets}

We examine the three data-sets proposed for data-to-text generation by
\cite{lebret-grangier-auli:2016:EMNLP2016}, \cite{wen2015semantically,wen2016multi}
and \cite{novikova2016analogue}.

\cite{lebret-grangier-auli:2016:EMNLP2016}'s data-set (\wikibio) 
focuses on biographies and associates Wikipedia
infobox with the first sentence of the corresponding article in
Wikipedia. Thus in this data-set, texts and data (infoboxes) were
authored by Wikipedia contributors. As the data-set is much larger than
the other data-sets and is not domain specific, we extract three
subsets of it for better comparison: two whose size is similar to the
other data-sets (\wikibio$_{16317}$, \wikibio$_{2647}$) and one which is domain
specific in that all biographies are about astronauts
(\wikibioastro ).

The other two data-sets were created manually with humans
providing text for dialogue acts in the case of
\cite{wen2015semantically,wen2016multi}'s data-sets
(\rnnlgLaptop , \rnnTV , \rnnlgHotel , \rnnlgRestaurant) and producing image descriptions in
the case of \cite{novikova2016analogue}'s data-set (\analogue ).

We also include a text-only corpus for comparison with the texts
contained in our three data-sets. This corpus (\gmb) consists of the
texts contained the Groningen Meaning Bank (Version 1.0.0,
\cite{basile2012developing}) and covers different genres (e.g., news,
jokes, fables).

\begin{table*}
\centering
\begin{scriptsize}
\begin{tabular}{lcccccccc}
Dataset & \size &  \OntComplexity & Entities$^\ddag$ & Attr.Values & MR Len. & MR Ptns & \MRInstances & PPxMR Inst.\\
\hline
\wikibio$_{16317}$& 16317&	2367&	16317 &	149484&	19.65 & 9990&	16317&		1\\
\wikibio$_{2647}$& 2647 & 1384& 2647 & 28375& 19.66 & 2083 & 2647 & 1\\
\wikibioastro& 615&	68&	615&	5290&	15.46 & 293&	615&	1\\\hline
\rnnlgLaptop&	13242&	34&	123&	451&	5.86 & 2068& 12527&	1.03(1/3)\\
\rnnTV&	7035&	30&	92&	300 & 5.79 & 1024&	6808&	1.01(1/6)\\
\rnnlgHotel&	5373&	22&	138 &	535&	2.66 & 112 &	940&	3.72(1/149)\\
\rnnlgRestaurant&	5192&	22&	223&	869 & 2.86 &	182&	1950&	1.82(1/101)\\\hline
\analogue&	1242&	16&	33&	117&	5.33 & 21&	77&	15.11(8/22)
\end{tabular} 
\end{scriptsize}
 
\caption{ \size\ is the number of instances in the data-set, (\OntComplexity) is the number 
of distinct attributes of the underlying ontology domain, (Entities) is the number of distinct topic entities described in each meaning representation,
(Attr.Values) is the number of attribute-value pairs, (MR Len.) is the average length of the input meaning representations computed as the number of attribute value pairs, (MR Ptns) is the number of distinct attribute combinations, namely MR patterns,  (\MRInstances) is the number of distinct 
meaning representations and (PPxMR Inst.) the average (min/max) number of paraphrases per meaning representation. $^\ddag$Note that the number of entities
is an approximation. We consider entities as distinct entities those given by the \textit{name} attributes and for the \rnn\ data-sets not all dialogue acts give entity descriptions. For instance, inform\_count or ?select .}\label{tab:data-setStats}
\end{table*}

\paragraph{Linguistic and Computational Adequacy}

Table~\ref{tab:data-setStats} gives some descriptive statistics for
each of these three data-sets. It shows marked differences in terms of
size ( \wikibio$_{16317}$ being the largest and \analogue\ the smallest),
number of distinct relations (from 16 for \analogue\ to 2367 for
\wikibio$_{16317}$ ) and average number of paraphrases (15.11 for
\analogue\ against 1 to 3.72 for the other two data-sets).  The number
of distinct meaning representations (semantic variability) also varies
widely (from 77 distinct MRs for the \analogue\ corpus to 12527 for
\rnnlgLaptop ).  Overall though, the main observation is that the
number of distinct attributes is relatively small and that the high
number of distinct meaning representations essentially stems from
various combinations of a restricted number of attributes.

\begin{figure*}[ht!]
  \begin{tabularx}{\linewidth}[t]{*{2}X}
    \begin{tabular}[l]{l}
{\scriptsize      
\begin{tabular}{lccccc}
Dataset &  Tokens & Types  & LS & MSTTR \\
\hline
\wikibio$_{16317}$ &377048	&36712		&0.92&	{\bf 0.82} \\
\wikibio$_{2647}$& 62321&	10391&		{\bf 0.85} &	{\bf 0.82}\\
\wikibioastro& 14720&	2335&		0.81&	0.8 \\
\rnnlgLaptop&	295492&	1757&		0.46&	0.74 \\
\rnnTV&	141606&	1171&		0.48&	0.71 \\
\rnnlgHotel&	48982&	967&		0.43&	0.59 \\
\rnnlgRestaurant&	45791&	1187&	0.43&	0.62 \\
\analogue&	20924&	598&	0.47&	0.56 \\
\hline
\hline
\gmb & 75927	&7791	&0.75	&0.81 \\
\end{tabular} 
}\end{tabular}
    &
    \begin{tabular}[l]{l}
    \hspace{-1.5em}
 \includegraphics[scale=0.4]{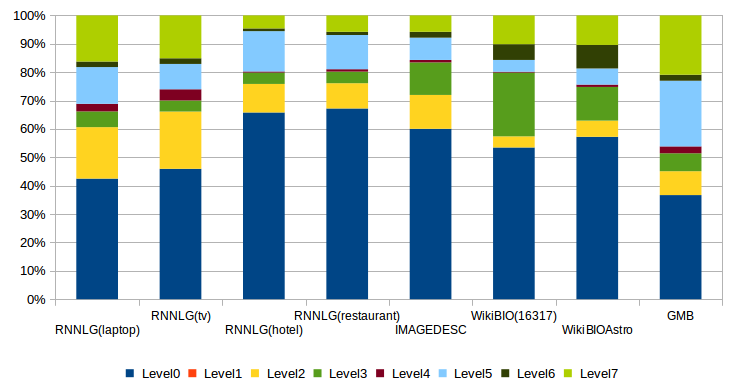}
    \end{tabular} \tabularnewline
    
\captionof{table}{Lexical Sophistication (LS) and Mean Segmental Type-Token Ratio (MSTTR). }\label{tab:lex}
    
    &
\captionof{figure}{Syntactic complexity D-Level sentence distribution.} \label{fig:syntax}
    \tabularnewline
\end{tabularx}

\end{figure*}

\paragraph{Lexical Richness} To assess the extent to which the three data-sets support the training of lexically rich surface realisers, we used the Lexical Complexity Analyser developed by \cite{lu2012relationship}, a system designed to automate the measurement of various dimensions of lexical richness. These include lexical sophistication (LS) and  mean segmental type-token ratio (MSTTR). Table~\ref{tab:lex} summarises the results.

\textit{Type-token ratio (TTR)} is a measure of diversity defined as
the ratio of the number of word types to the number of words in a
text. To address the fact that this ratio tends to decrease with the
size of the corpus, Mean segmental TTR (MSTTR) is computed by dividing
the corpus into successive segments of a given length and then
calculating the average TTR of all segments. 

Overall, the \wikibio\ data-set, even when restricted to a single type
of entity (namely, astronauts) has a higher MSTTR. This higher lexical
variation is probably due to the fact that this data-set also
has the highest number of attributes
(cf. Table~\ref{tab:data-setStats}): more attributes brings more
diversity and thus better lexical range. And indeed, there is a
positive correlation between the number of attributes in the data-set
and MSTTR (Spearman’s rank correlation coefficient rho = +0.385).

\textit{Lexical sophistication}, also known as lexical rareness,
measures the proportion of relatively unusual or advanced word types
in the text. In practice, LS is the proportion of lexical word types
(lemma) which are not in the list of 2,000 most frequent words
generated from the British National Corpus. Again the
\wikibio\ data-set has a markedly higher level of lexical
sophistication than the other data-sets. This might be because the
\wikibio\ text are not crowd-sourced but edited independently of input
data thereby leaving more freedom to the authors to include additional
information. It may also result from the fact that the
\wikibio\ data-set, even though it is restricted to biographies, covers
a much more varied set of domains than the other data-sets as people's
lives may be very diverse and consequently, a more varied range of
topics may be mentioned than in a domain restricted
data-set.



\paragraph{Syntactic Diversity}
To support the training of surface realisers with wide syntactic
coverage, a benchmark needs to show a balanced distribution of the
various syntactic phenomena present in the target language. To compare
the syntactic coverage of the three data-sets, we use the system for
automatic measurement of syntactic complexity developed by
\cite{lu2010automatic}. Briefly, this system decomposes parse
trees\footnote{Parses are obtained using Collins' constituent parser
  \cite{collins99}.} into component sub-trees and scores each of these
sub-trees based on the type of the syntactic constructions detected in
it using a set of heuristics. Sentences are then assigned to a
syntactic level based on the scores assigned to the sub-trees it
contains as follows. If all sub-trees found in that sentence are
assigned to level zero, the sentence is assigned to level 0; if one
and only one non-zero level is assigned to one or more sub-trees, the
sentence is assigned to that non-zero level; if two or more different
non-zero scores are assigned to two or more of the sub-trees, the
sentence is assigned to level 7.  When evaluated against a gold
standard of 500 sentences\footnote{The gold sentences were
  independently rated by two annotators yielding a very high
  inter-annotator agreement (kappa = 0.9108).}, the system achieves a
precision of 93.2\%, a recall of 93.2\% and an F-Score of 93.2\%
\cite{lu2010automatic}.

The system uses the revised D-Level Scale proposed by
\cite{covington2006complex} which consists of the following eight levels: (0)
simple sentences, including questions (1) infinitive or -ing
complement with subject control; (2) conjoined noun phrases in subject
position; conjunctions of sentences, of verbal, adjectival, or
adverbial construction; (3) relative or appositional clause modifying
the object of the main verb; nominalization in object position; finite
clause as object of main verb; subject extraposition; (4) subordinate
clauses; comparatives; (5) nonfinite clauses in adjunct positions; (6)
relative or appositional clause modifying subject of main verb;
embedded clause serving as subject of main verb; nominalization
serving as subject of main verb; (7) more than one level of embedding
in a single sentence.

Figure~\ref{fig:syntax} summarises the results for the various data-sets.
There are several noticeable issues.

First, the proportion of simple texts (Level 0) is very high across
the board (42\% to 68\%). In fact, in all data-sets but two,
\textit{more than half of the sentences are of level 0 (simple
  sentences)}. In comparison, only 35\% of the GMB corpus 
sentences are of level 0.

Second, levels 1, 4 and to a lesser extent level 3, are absent or
almost absent from the data-sets. We conjecture that this is due to
the shape and type of the input data. Infinitival clauses with subject
control (level 1) and comparatives (level 4) involve coreferential
links and relations between entities which are absent from the simple
frames comprising the input data. Similarly, non finite complements
with their own subject (e.g., \nl{``John saw Mary leaving the room''},
Level 3) and relative clauses modifying the object of the main verb
(e.g., \nl{``The man scolded the boy who stole the bicycle''}, Level 3)
requires data where the object of a
literal is the subject of some other literal i.e. data  of the
form $P_x (x) R_1 (x \; y) P_y (y) R_2 (y \; z) P_z (z)$. In most cases
however, the input data consists of sets of literals predicating facts
about a single entity i.e., literals of the form $P_x (x) R_1 (x \; y)
P_y (y) R_2 (y \; z) P_z (z)$ where $x$ is the entity to be described.

Third, the choice of domain seems to impact syntactic complexity.
Thus the two data-sets in the restaurant domain, although crowd-sourced
using different methods, have a similar distribution. Conversely, the
various \rnn\ data-sets, although developed using the same method,
display different distributions. The mean D-level also seems to be
impacted by the length of the meaning representation. Thus, in the
\rnn\ data-sets where the length of the meaning representations (MR)
varies (cf. Table~\ref{tab:data-setStats}), the Spearman's rank
correlation between MR length and D-level is +0.9 with $p$=0.037.

Fourth, data-sets may be more or less varied in terms of syntactic
complexity. It is in particular noticeable that, for the
\wikibio\ data-set, three levels (1, 3 and 7) covers 84\% of the
cases. This restricted variety points to stereotyped text with
repetitive syntactic structure. Indeed, in \wikibio , the texts
consist of the first sentence of biographic Wikipedia articles which
typically are of the form \nl{``W (date of birth - date of death) was
  P''.} where $P$ usually is an arbitrarily complex predicate
potentially involving relative clauses modifying the object of main
verb (Level 3) and coordination (Level 7).

\paragraph{Semantic Adequacy} A surface realiser ought to express all and only the content captured in the input data. We therefore investigate to what extent data/text pairs of each data-sets match this requirement by manually examining 50 input/output pairs randomly extracted from each data-set. A data/text pair was considered
 a correct match if all slot/values pairs present in the input were
 verbalised in the text. Conversely, it was annotated as
 ``Additional'' if the text contained information not present in the
 data and as ``Missing'' if the data contained information not present
 in the text. Each pair was independently rated by two annotators
 resulting in a kappa score ranging between 0.566 and 0.691 depending
 on the data-set.
 The results shown in Table~\ref{tab:semAnnotation} highlight some important
 differences. While the \rnn\ data-sets have a high percentage of
 correct entries (82\% to 94\%), the \analogue\ data-set is less
 precise (44\% of correct matches). The \wikibio\ data-sets does not
 contain a single example where data and text coincides. These
 differences can be traced back to the way in which each resource was
 created. In most of the cases (70\%) annotators indicated both
 missing and additional content.

The \wikibio\ data-set is created automatically from Wikipedia
infoboxes and articles while semantic adequacy is not checked for. 
As such, this data-set is ill-suited for training precise surface
realisers. Moreover, its texts contain both missing and
additional information. Due to this mismatch, it cannot be used to train joint models for
content selection and surface realisation with standard neural
techniques \cite{sutskever2014sequence}.

In the \analogue\ data-set, the texts are created from images using
crowd-sourcing. It seems that this method, while enhancing variety,
makes it easier for the crowd-workers to omit some information thereby resulting in several cases where the text fails to express all the information contained in the input data.
  
\begin{table}
\centering
{\scriptsize
\begin{tabular}{lccc}
\hline
& M & A & C \\
\rnnlgLaptop & 16\% & 2\% & 82\% \\
\rnnTV & 12\% & 4\%  & 84\% \\ 
\rnnlgHotel & 0 & 6\% & 94\% \\ 
\rnnlgRestaurant & 0 & 6\% &94\%  \\
\analogue & 50\% & 6\% & 44\% \\ 
\hline
& M & A & MA \\
\wikibioastro & 30\% & 0 & 70\% \\
\hline
\end{tabular}  
}
\caption{Match between Text and Data. M: Missing content in the text, A: Additional content in the text, MA: both additional and missing, C:correct.}\label{tab:semAnnotation}
  \end{table}

\section{Conclusion}

Using statistics, metrics and manual evaluation, we compared three
surface realisation benchmarks.  Unsurprisingly,
our analysis shows that existing data-sets have restrictions (large
proportion of simple sentences, absence of certain syntactic
constructions, limited syntactic and semantic diversity, restricted
number of attributes and of distinct inputs, stereotyped text,
etc.). On a more positive note, it also suggests several key aspects
to take into account when constructing a data-to-text data-set for the
development, evaluation and comparison of linguistically sophisticated
surface realisers. \textit{Lexical richness} can be enhanced by
including data from different domains, using a large number of
distinct attributes and ensuring that the total number of distinct
inputs is high. Wide and balanced \textit{syntactic coverage} is
difficult to ensure and probably requires input data of various size
and shape, stemming from different domains (biographic text for
instance, has limited syntactic variability). \textit{Semantic
  adequacy} can be achieved almost perfectly using crowd-sourcing which
also facilitates the inclusion of paraphrases.

\bibliography{d2tbenchmarks}

\begin{thebibliography}{}

\bibitem[\protect\citename{Basile \bgroup et al.\egroup
  }2012]{basile2012developing}
Valerio Basile, Johan Bos, Kilian Evang, and Noortje Venhuizen.
\newblock 2012.
\newblock Developing a large semantically annotated corpus.
\newblock In {\em LREC}, volume~12, pages 3196--3200.

\bibitem[\protect\citename{Collins}1999]{collins99}
M.~Collins.
\newblock 1999.
\newblock {\em Head-Driven Statistical Models for Natural Language Parsing}.
\newblock {Ph.D.} thesis, University of Pennsylvania, Philadelphia, PA.

\bibitem[\protect\citename{Covington \bgroup et al.\egroup
  }2006]{covington2006complex}
Michael~A Covington, Congzhou He, Cati Brown, Lorina Naci, and John Brown.
\newblock 2006.
\newblock How complex is that sentence? a proposed revision of the rosenberg
  and abbeduto d-level scale.

\bibitem[\protect\citename{Lebret \bgroup et al.\egroup
  }2016]{lebret-grangier-auli:2016:EMNLP2016}
R\'{e}mi Lebret, David Grangier, and Michael Auli.
\newblock 2016.
\newblock Neural text generation from structured data with application to the
  biography domain.
\newblock In {\em Proceedings of the 2016 Conference on Empirical Methods in
  Natural Language Processing}, pages 1203--1213, Austin, Texas, November.
  Association for Computational Linguistics.

\bibitem[\protect\citename{Lu}2010]{lu2010automatic}
Xiaofei Lu.
\newblock 2010.
\newblock Automatic analysis of syntactic complexity in second language
  writing.
\newblock {\em International Journal of Corpus Linguistics}, 15(4):474--496.

\bibitem[\protect\citename{Lu}2012]{lu2012relationship}
Xiaofei Lu.
\newblock 2012.
\newblock The relationship of lexical richness to the quality of esl
  learners’ oral narratives.
\newblock {\em The Modern Language Journal}, 96(2):190--208.

\bibitem[\protect\citename{Novikova and Rieser}2016]{novikova2016analogue}
Jekaterina Novikova and Verena Rieser.
\newblock 2016.
\newblock The analogue challenge: Non aligned language generation.
\newblock In {\em The 9th International Natural Language Generation
  conference}, page 168.

\bibitem[\protect\citename{Novikova \bgroup et al.\egroup
  }2016]{novikova-lemon-rieser:2016:INLG}
Jekaterina Novikova, Oliver Lemon, and Verena Rieser.
\newblock 2016.
\newblock Crowd-sourcing nlg data: Pictures elicit better data.
\newblock In {\em Proceedings of the 9th International Natural Language
  Generation conference}, pages 265--273, Edinburgh, UK, September 5-8.
  Association for Computational Linguistics.

\bibitem[\protect\citename{Sutskever \bgroup et al.\egroup
  }2014]{sutskever2014sequence}
Ilya Sutskever, Oriol Vinyals, and Quoc~V Le.
\newblock 2014.
\newblock Sequence to sequence learning with neural networks.
\newblock In {\em Advances in neural information processing systems}, pages
  3104--3112.

\bibitem[\protect\citename{Wen \bgroup et al.\egroup
  }2015]{wen2015semantically}
Tsung-Hsien Wen, Milica Gasic, Nikola Mrksic, Pei-Hao Su, David Vandyke, and
  Steve Young.
\newblock 2015.
\newblock Semantically conditioned lstm-based natural language generation for
  spoken dialogue systems.
\newblock {\em arXiv preprint arXiv:1508.01745}.

\bibitem[\protect\citename{Wen \bgroup et al.\egroup }2016]{wen2016multi}
Tsung-Hsien Wen, Milica Gasic, Nikola Mrksic, Lina~M Rojas-Barahona, Pei-Hao
  Su, David Vandyke, and Steve Young.
\newblock 2016.
\newblock Multi-domain neural network language generation for spoken dialogue
  systems.
\newblock {\em arXiv preprint arXiv:1603.01232}.

\end{thebibliography}
\bibliographystyle{naaclhlt2016}

\end{document}